\documentclass{article} 
\usepackage[final]{colm2025_conference}

\usepackage{microtype}
\usepackage{hyperref}
\usepackage{url}
\usepackage{booktabs}
\usepackage{graphicx}
\usepackage{lineno}

\definecolor{darkblue}{rgb}{0, 0, 0.5}
\hypersetup{colorlinks=true, citecolor=darkblue, linkcolor=darkblue, urlcolor=darkblue}

\title{ACT: Bridging the Gap in Code Translation through Synthetic Data Generation \& Adaptive Training}

\author{Shreya Saxena, Siva Prasad, Zishan Ahmad, Vishal Vaddina
\\
Phi Labs, Quantiphi Analytics \\
\texttt{\{shreya.saxena, siva.prasad, zishan.ahmad, vishal.vaddina\}@quantiphi.com} \\
}

\begin{document}


\maketitle

\begin{abstract}
Code translation is a crucial process in software development and migration projects, enabling interoperability between different programming languages and enhancing software adaptability and thus longevity. Traditional automated translation methods rely heavily on handcrafted transformation rules, which often lack flexibility and scalability. Meanwhile, advanced language models present promising alternatives but are often limited by proprietary, API-based implementations that raise concerns over data security and reliance. In this paper, we present Auto-Train for Code Translation (ACT), an innovative framework that aims to improve code translation capabilities by enabling in-house finetuning of open-source Large Language Models (LLMs). ACT’s automated pipeline significantly boosts the performance of these models, narrowing the gap between open-source accessibility and the high performance of closed-source solutions. Central to ACT is its synthetic data generation module, which builds extensive, high-quality datasets from initial code samples, incorporating unit tests to ensure functional accuracy and diversity. ACT’s evaluation framework incorporates execution-level checks, offering a comprehensive assessment of translation quality. A key feature in ACT is its controller module, which manages the entire pipeline by dynamically adjusting hyperparameters, orchestrating iterative data generation, and finetuning based on real-time evaluations. This enables ACT to intelligently optimize when to continue training, generate additional targeted training data, or stop the process. Our results demonstrate that ACT consistently enhances the effectiveness of open-source models, offering businesses and developers a secure and reliable alternative. Additionally, applying our data generation pipeline to industry-scale migration projects has led to a notable increase in developer acceleration.
\end{abstract}

\section{Introduction}
\begin{figure*}[ht]
    \centering
\includegraphics[width=\linewidth,trim=0.3cm 0.4cm 0.2cm 0.2cm,clip]{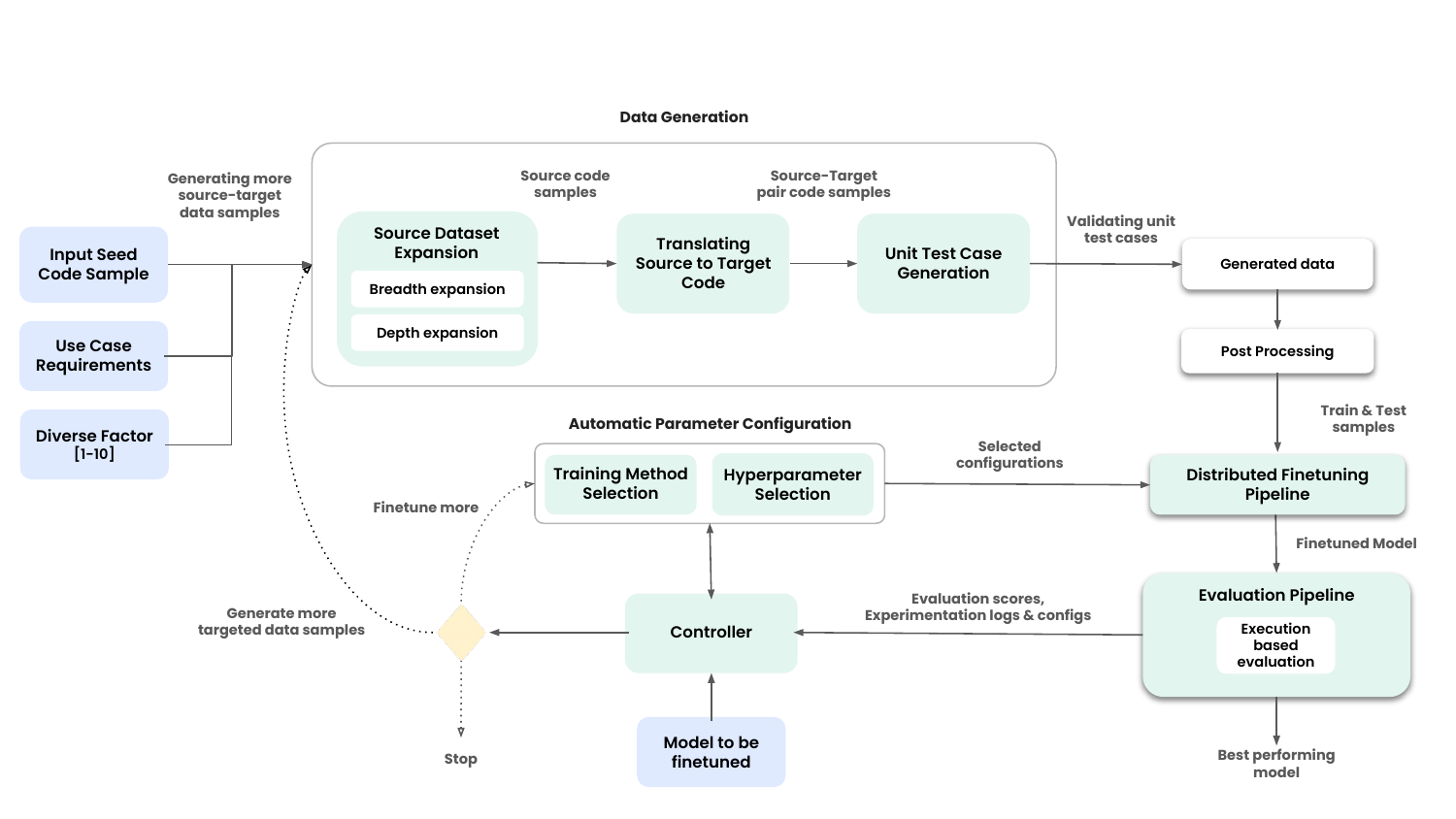}
    \caption{\label{fig:framework}Overview of the ACT framework, illustrating the workflow with a focus on the Controller's role in managing iterative processes like data generation and finetuning }
\end{figure*}

In recent years, the rapid evolution of software development practices has necessitated seamless code translation across diverse programming languages. Traditional code translation methods often relied on manually created rules to transform code, were labor intensive and often produced suboptimal translations \cite{roziere2020unsupervised, mukherjee2011automatic}. Statistical machine translation techniques for code translation \cite{karaivanov2014phrase} such as semSMT \cite{nguyen2014migrating} and others using abstract syntax tree (AST) \cite{phan2020statistical} improved translation quality, but the advent of large language models (LLM) significantly advanced the quality of code translation systems \cite{weisz2022better}. While open-source language models show they often lag behind proprietary solutions \cite{roziere2023code, hui2024qwen2}. Proprietary, closed-source models deliver superior results but pose concerns about data security, control, and dependency on third-party services. Finetuning open-source models is a viable solution, but limited access to high-quality training data remains a critical challenge. Recent research has explored synthetic data generation as a means to improve fine-tuning for various tasks, including code translation \cite{xu2023wizardlm, lu2024mathgenie}. However, effectively generating diverse, functionally accurate synthetic data remains an open problem.

Building on these advancements, we introduce Auto-Train for Code Translation (ACT), a novel framework that facilitates in-house finetuning of open-source LLMs for effective and secure code translation. Central to ACT is its automatic synthetic data generation module, which generates extensive, high-quality datasets from an initial set of code samples provided, complete with unit tests to validate functional accuracy. Through iterative data generation and model finetuning, ACT optimizes translation capabilities with minimal data overhead, forming a robust solution tailored to specific project needs.

\section{Methodology}
The Auto Train framework for code translation consists of four primary components: (i). Data Generation, (ii). Finetuning, (iii). Evaluation, and (iv). Controller. Each component plays a crucial role in automating the process of generating synthetic code data, finetuning models, evaluating their performance, and managing iterative workflows. This section provides a comprehensive overview of the framework, as illustrated in Figure 1.

\subsection{Data Generation}
The data generation component, as shown in Figure 2, serves as the cornerstone of the Auto Train framework for code translation, as it produces high-quality training and validation data essential for finetuning an accurate code translation model. This section outlines the steps involved in generating both source and target language code samples, along with unit test cases essential for verifying the quality of the generated code making use of LLMs.
\footnote{Refer to Appendix for a working example.}.
\begin{figure*}[ht]
    \centering
    \includegraphics[width=\linewidth,trim=0.3cm 0.4cm 0.2cm 0.2cm,clip]{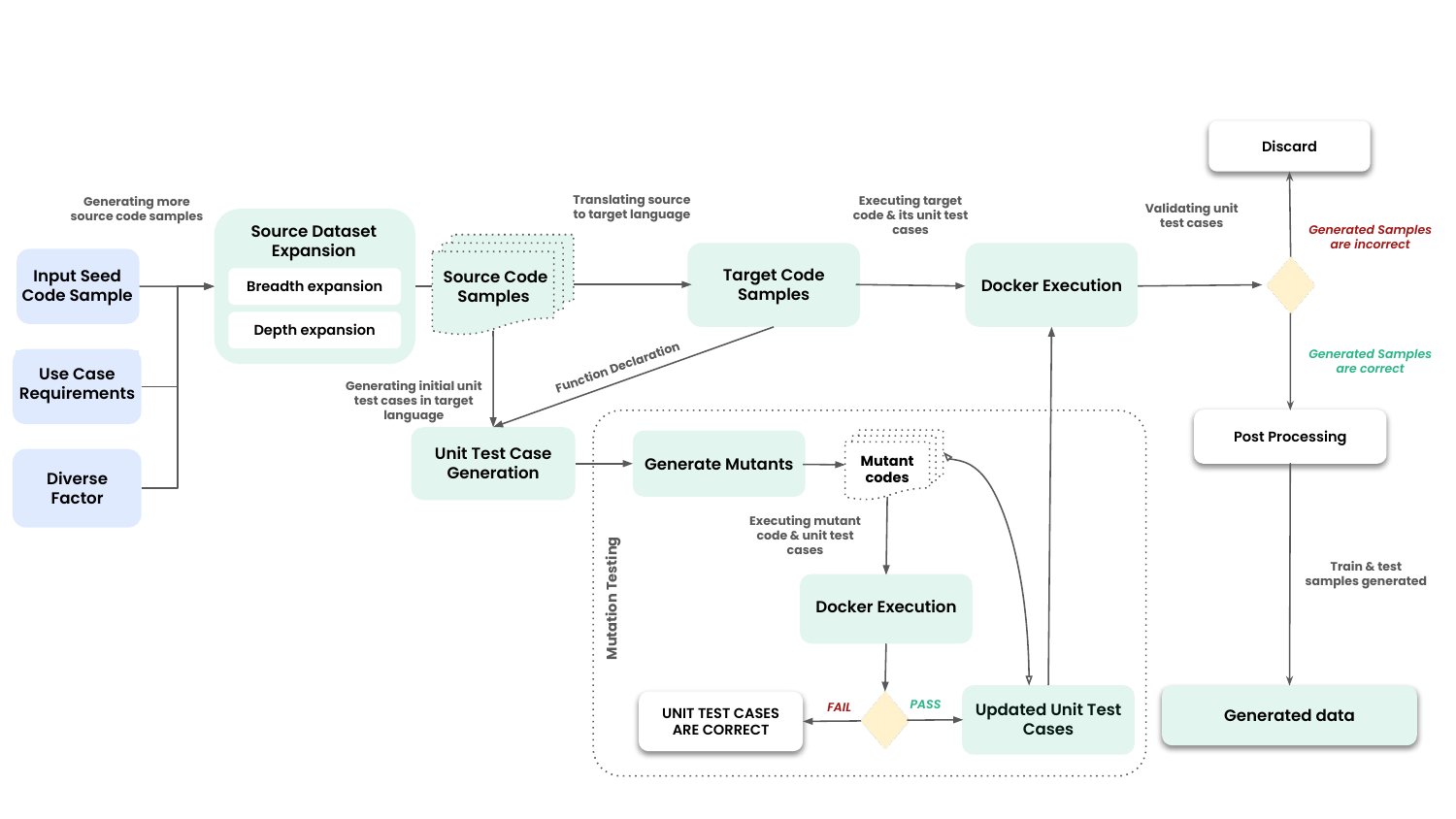}
    \caption{Workflow of the Data Generation Process - includes dataset expansion, translation, unit testcase generation, data validation through execution, and processing of train and test splits for further model finetuning.
}
\end{figure*}

\subsubsection{Dataset Expansion}
The process begins with seed code samples in the source language, which users provide along with a "diverse factor" and specific use case requirements. These initial samples serve as a basis for generating synthetic source language code samples that enrich the training dataset. The framework captures a wide variety of coding styles and structures to enhance the robustness of the model. Two key strategies are employed for dataset expansion \cite{xu2023wizardlm}:

\textbf{Breadth-Based Expansion}: This approach focuses on diversifying the generated code by altering core objectives, allowing for a wider variety of code samples to be produced. For example, a simple seed input that checks whether a number is even or odd could be extended to check if a number is a palindrome, thereby introducing functional diversity.

\textbf{Depth-Based Expansion}: This strategy adds complexity to the seed samples by incorporating constraints or modifying reasoning steps. The framework employs specific prompts to achieve this: \\
\textit{i) Constraints Prompt:} Introduce new requirements or conditions. \\
\textit{ii) Deepen Prompt:} Alter time or space complexity requirements. \\
\textit{iii) Concretizing Prompt:} Replace an existing requirement with a specific condition. \\
\textit{iv) Reasoning Prompt:} Modify the logical flow and reasoning steps.

 The diverse factor, which ranges from 1 to 10, controls how much the generated code deviates from the input seed samples. A higher diverse factor indicates a stronger emphasis on breadth-based expansion over depth-based expansion and vice versa. Additionally, users can specify particular use case requirements, such as requesting lower time complexity or the inclusion of docstrings. Depending on the diverse factor and user requirements, the source language seed code samples are expanded using both depth-based and breadth-based expansion methods with one or more LLMs to generate synthetic data. Employing multiple LLMs further enhances variety in the generated synthetic data.

\subsubsection{Translating to Target Language} Once the source language code samples are generated, the next step involves translating these samples into the target programming language. For translating large code files, the approach considered is dividing the entire code into sub-segments, translating each sub-segment individually, and merging back them using LLMs. Depending on the language, these segments could be `classes', `functions', or `transformation steps' (in the case of ETL pipelines).


\subsubsection{Unit Test Case Generation} Unit test cases are generated for each translated code sample to ensure semantic and syntactic accuracy, validating the expected behavior when executed. The initial test cases are derived from the original source code, focusing on the function declarations of the translated code rather than the translated code itself.  This approach helps avoid translation biases introduced during translation and allows testing of the core logic of the source code. 

To enhance the reliability and coverage of the initial unit test cases generated, the framework employs mutation testing. Mutation testing \cite{dakhel2024effective} involves creating "mutants" buggy versions of the translated code-by introducing small modifications that could change the logic of the program. These modifications may involve tweaking conditional, arithmetic, assignment, relational, or logical operators. Our approach creates mutants by adding a single bug to the translated code, generating multiple mutants to test various code aspects and improve test coverage. The mutants are then tested against the initial unit cases. If a mutant passes all tests, it suggests that the tests may not comprehensively cover the translated code. In response, the unit tests are revised and refined to better detect these issues, ensuring a more robust test suite that can identify a wider range of potential flaws.


\subsubsection{Validation of Synthetic Data} 
The generated target code samples are validated by executing them against their unit test cases in Docker containers. This controlled setup ensures that all necessary libraries and dependencies are available for each code sample. The process is fully automated using a custom Docker Manager that handles container setup, library installation, and sequential code execution.

To begin, the appropriate Docker container for the target programming language is created. The translated code and its associated unit tests are organized appropriately, either in the same or separate files as required by the testing framework.  Additional processing may be required depending on the language;for example, in Go, the import statements must be defined only once at the top of the file. After ensuring that all necessary files are correctly configured, the required external libraries are installed. Te translated code is executed within the Docker container, and the unit tests are run to verify the correctness of the code. Only samples that pass all tests are retained for further finetuning and evaluation; those that fail are discarded. This ensures that the training data consists of well-tested, functional code, enhancing the model's quality. 

In subsequent iterative stages, additional targeted data samples are generated to address coverage or performance gaps. Failed samples return to the data generation stage for breadth-based expansion, optimizing the dataset for training success.

\subsection{Finetuning}
The finetuning component is dedicated to optimizing models for code translation tasks using the high-quality synthetic data produced during the data generation phase. The distributed training pipeline leverages Deepspeed \cite{rajbhandari2020zero} and Pytorch Lightning for faster and memory-efficient finetuning, incorporating techniques like Mixed-Precision Training, Quantized Training, Gradient Checkpointing and Accumulation, Flash Attention, and Deepspeed Zero strategies along with CPU offloading. The finetuning process is driven by the automatic parameter configuration module.

\textbf{Automatic Parameter Configuration}: The framework autonomously selects the necessary training hyperparameters and configurations required for finetuning the model using an LLM. The users don't need any technical knowledge to finetune LLMs such as the finetuning strategies, number of epochs, batch size, or other parameters. This automatic configuration occurs through two key steps:\\
   \textit{ i) Training Strategy Selection:} Depending on the evaluation scores of the base model on the test data and the size of the training dataset, the framework determines which finetuning strategy to employ- Full Supervised Finetuning (SFT) or Low-Rank Adaptation (LoRA) \cite{hu2021lora}. If the evaluation scores are low, indicating the model's limited familiarity with the code translation task, full finetuning could be chosen. Conversely, if the scores are good, LoRA finetuning can be utilized. \\
   \textit{ ii) Training Configuration Selection:} This step automatically adjusts various training parameters such as the number of epochs, training and validation batch sizes per GPU, gradient accumulation size, Deepspeed strategy with/without CPU offloading, and learning rate. These decisions are based on factors such as dataset size, model size, and available GPU configuration. Additionally, LoRA-specific parameters like rank ($r$) and $\alpha$ values are also automatically setup in this step.


The finetuning process follows an iterative approach, with evaluations after each training stage to assess model performance. Decisions to continue training, adjust parameters, generate more data, or finalize the experiment with the best-performing model depend on evaluation scores and trends in training and validation loss curves.

\subsection{Evaluation}
Model evaluation is performed after each iteration of finetuning to track and measure the model's performance, with the best-performing model being tracked throughout the process. The evaluation ensures a comprehensive review of the functional correctness of the translated code. The unit test cases are assumed to be comprehensive and sufficient due to mutation testing that tests every aspect of the translated code.  

    \textbf{Execution-Level Evaluation}: The generated target code is executed within a secure Docker container against the generated unit test cases. This process computes \textit{pass@k} values, providing a quantitative measure of model performance. Given a source code, the model generates the top k most probable outputs (i.e., translated code samples). These outputs are then tested against a set of unit tests designed to validate functional correctness. The \textit{pass@k} metric is defined as the probability that at least one of the k-generated outputs passes predefined test cases. 
    

    

The evaluation results are fed back to the Controller, which makes informed decisions about whether to continue finetuning, generate additional data, or stop the process, thereby ensuring iterative improvement in model performance.

\subsection{Controller}
The Controller orchestrates the Auto-Train framework, managing the interplay between data generation, finetuning, and evaluation across multiple stages. At each stage, it dynamically decides whether to:- i) Continue finetuning the model if the model shows consistent improvement. ii) Generate additional synthetic data if the evaluation suggests that data diversity or quality is a limiting factor in model improvement. iii) Terminate the finetuning process if further iterations lead to diminishing returns. This iterative optimization process, powered by an LLM-based decision mechanism, optimizes each stage based on prior results, maximizing model performance while minimizing unnecessary training cycles, time, compute, and cost, leading to a more efficient finetuning pipeline.

A key innovation of the Controller is its ability to adaptively determine the optimal data size at each finetuning stage. Instead of generating large datasets upfront, which can lead to unnecessary computation, the Controller ensures that only the necessary volume of high-quality synthetic data is generated. At each stage, the Controller evaluates model performance based on the training and validation loss values and evaluation scores of the base model, finetuned model of the current stage, and previous stages to guide decisions:-

\textbf{Targeted Data Generation}: Synthetic examples are iteratively crafted to address failed cases identified in prior stages, ensuring context-aware data augmentation.

\textbf{Adaptive Finetuning}:  Hyperparameters—epochs, learning rate, batch size, and gradient accumulation steps—are dynamically adjusted every stage based on loss and evaluation trends. This adaptive approach prevents overfitting while accelerating convergence.

By leveraging data-driven decisions, LLM optimization, and historical tracking, the Controller enhances training efficiency, ensuring Auto-Train remains scalable and effective for large-scale code translation and generative AI tasks.

\section{Experimental Setup}
In this paper, we report evaluations on two open-source models, Qwen2.5-Coder-7B-Instruct \cite{hui2024qwen2} and DeepSeek-Coder-V2-Instruct \cite{zhu2024deepseek}, applied to code translation tasks for Java-to-Go and C++-to-Rust conversions. The primary goal was to assess our finetuning approach by comparing the performance of the finetuned models against that of the base models on these translation tasks.

The experiments began with a seed dataset of 150 source-target code pairs from the HumanEval dataset \cite{chen2021evaluating}, which was then systematically expanded to provide the models with a larger and more diverse code set. For each seed sample, one breadth-expanded and four depth-expanded variations (with a diversity factor set to 1) were generated. GPT-4o and Claude-3.5-Sonnet were used for the data generation process. A subset of 30 generated training samples per translation task was manually evaluated by three software developers with relevant programming expertise. Of these, 15 were marked as correct and 15 as incorrect by the data generation pipeline during the execution-based evaluation. The correct samples were reviewed to confirm the accuracy of their validation, while the incorrect samples were analyzed to verify whether they were properly identified as failures and to investigate the reasons for their failure. They assessed the translated code for correctness and unit test cases for coverage and accuracy, scoring them on a scale of 0 to 2, where 0 indicates completely incorrect and unaligned, 1 represents partial alignment, and 2 signifies complete correctness and alignment. The mean score of all the samples is shown in Table 1.


Hyperparameters critical to finetuning-such as LoRA-specific parameters (with $r$ values of 16 or 32 and $\alpha$ set to 64), epochs (ranging from 2 to 5), learning rates (from 3e-5 to 1e-6), and batch sizes (1 to 4)-were optimized based on model specifications, GPU constraints, and dataset size by the automatic training configuration module with the model loaded in 8-bit by default. Training was conducted in stages using a single 80-GB A100 GPU, with 450 samples on average for initial training and adding an additional 100 samples for each subsequent stage, along with a validation dataset of 50 samples. The controller module monitored the training process, deciding whether to continue or halt based on performance improvements, with a cap limit of four stages.

\begin{table}[h]
    \centering
    \renewcommand{\arraystretch}{1.2}
    \begin{tabular}{lcc|cc}
        \hline
        \textbf{Task} & \multicolumn{2}{c}{\textbf{C++ → Rust}} & \multicolumn{2}{c}{\textbf{Java → Go}} \\
        \cline{2-5}
        & \textbf{Unit Test Case} & \textbf{Translated Code} & \textbf{Unit Test Case} & \textbf{Translated Code} \\
        \hline
        \textbf{Correct}   & 1.93  & 2.00  & 1.80  & 2.00  \\
        \textbf{Incorrect} & 1.47  & 1.60  & 1.47  & 1.53  \\
        \hline
    \end{tabular}
    \caption{Human evaluation results of the training samples by the data generation pipeline. The scores reflect the evaluation of both translated code and unit test cases, with ratings (on a scale of 0-2) based on correctness and coverage.}
    \label{tab:human_eval}
\end{table}

\section{Results}
The human evaluation results, as shown in Table 1, highlight the effectiveness of the data generation pipeline in producing high-quality translated code and unit test cases. The correctly filtered samples received high ratings, with translated code achieving a perfect score and unit test cases scoring nearly perfect, indicating accurate translations and minimal loss in test coverage. In contrast, the incorrectly filtered samples received comparatively lower scores, reflecting partial coverage of unit test cases and some errors in the translated code. The errors in these samples were primarily due to issues such as syntax errors and logic implementation flaws. This confirms that the pipeline correctly identified these samples as faulty. Overall, the data generation pipeline effectively filters, validates, and produces high-quality training data.

The experimental results confirm that the ACT finetuning process significantly improves code translation accuracy for both Qwen2.5-Coder-7B-Instruct and DeepSeek-Coder-V2-Instruct models. As detailed in Table 1, both ACT-Qwen2.5-Coder-7B-Instruct and ACT-DeepSeek-Coder-V2-Instruct outperform their base models in the Java-to-Go translation task, with DeepSeek demonstrating notable adaptability to finetuning. In the C++-to-Rust translation task, ACT-DeepSeek-Coder-V2-Instruct achieves substantial improvements, while ACT-Qwen2.5-Coder-7B-Instruct shows a slight decline (\textit{pass@1}), hinting at a balance between precision and generalization challenges for Qwen-based models. However, the finetuned model still outperforms the base by a big marking when evaluated for \textit{pass@5}.  Overall, both models demonstrate greater effectiveness in adapting to Java-to-Go translation compared to C++-to-Rust translation.

\begin{table}[t]
\tiny
\caption{Execution-based metric comparing accuracy between base model \& ACT finetuned model}
\label{tab:my-table}
\resizebox{\columnwidth}{!}{%
\begin{tabular}
{|l|cc|lc|}
\hline
\multicolumn{1}{|c|}{}        & \multicolumn{2}{c|}{\textbf{Java to Go}}                & \multicolumn{2}{c|}{\textbf{ C++ to Rust}}                             \\ \cline{2-5} 
\multicolumn{1}{|c|}{{\textbf{ Models}}} &
  \multicolumn{1}{l|}{\textbf{ \textit{pass@1}}} &
  \multicolumn{1}{l|}{\textbf{ \textit{pass@5}}} &
  \multicolumn{1}{l|}{\textbf{ \textit{pass@1}}} &
  \multicolumn{1}{l|}{\textbf{ \textit{pass@5}}} \\ \hline
 Qwen2.5-Coder-7B-Instruct     & \multicolumn{1}{c|} {0.4461}          &  0.5904          & \multicolumn{1}{c|}{\textbf{0.4184}}               &  0.5646          \\
 ACT-Qwen2.5-Coder-7B-Instruct & \multicolumn{1}{c|}{\textbf{ 0.4865}} & \textbf{ 0.6642} & \multicolumn{1}{c|}{{ 0.3950}} & \textbf{ 0.6192} \\ \hline
 DeepSeek-Coder-V2-Instruct &
  \multicolumn{1}{c|}{ 0.5348} &
   0.6048 &
  \multicolumn{1}{c|}{{ 0.4500}} &
   0.5612 \\
 ACT-DeepSeek-Coder-V2-Instruct &
  \multicolumn{1}{c|}{\textbf{ 0.6248}} &
  \textbf{ 0.7019} &
  \multicolumn{1}{c|}{\textbf{ 0.5653}} &
 \textbf{ 0.6328} \\ \hline
\end{tabular}%
}
\end{table}

\begin{figure}[ht]
    \centering
    \includegraphics[width=0.4\linewidth,trim=0.3cm 0.4cm 0.2cm 0.2cm,clip]{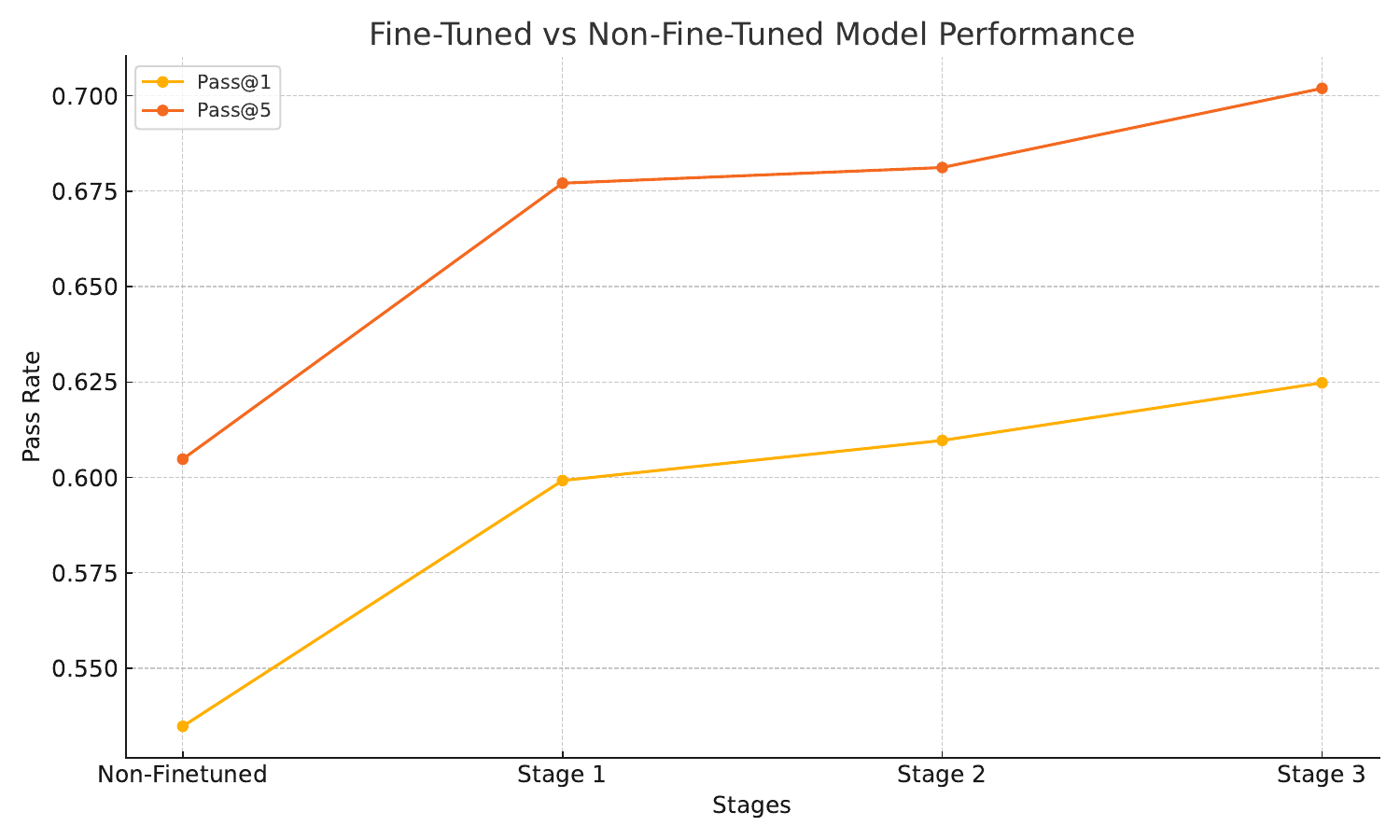}
    \caption{\label{fig:graph}Execution-based evaluation scores (\textit{pass@1} and \textit{pass@5}) of DeepSeek-Coder-V2-Instruct for Java-Go translation across finetuning stages, showcasing iterative improvements
}
\end{figure}

The experiments highlight the effectiveness of the iterative finetuning approach, with significant improvements in execution-based evaluation scores observed at each finetuning stage. As depicted in Figure \ref{fig:graph}, Stage 1 saw DeepSeek-Coder-V2-Instruct generate 500 Java-Go samples from 150 seed samples and undergo finetuning for three epochs. In Stage 2, the controller determined that two additional epochs would yield further improvements, resulting in a modest increase in evaluation scores. In Stage 3, the controller generated 100 targeted samples from previous failures and finetuned the model for two more epochs, leading to a substantial performance boost. Improvement was consistently observed across all experiments. In another scenario, the controller terminated the process after two stages when no further gains were anticipated, underscoring its ability to make informed and adaptive decisions for optimal finetuning and model performance.

In conclusion, the controller-driven, data generation, and finetuning framework effectively enhances code translation tasks. By dynamically adjusting fine-tuning and generating targeted data, it ensures continuous model improvement.


\section{Conclusion \& Future Work}
In this paper, we introduced Auto-Train for Code Translation (ACT), an advanced framework designed to enhance open-source language models for code translation tasks, offering a secure and high-performing alternative to proprietary solutions. ACT tackles significant challenges in code translation, including limited training data and the complexities of cross-language code validation. Our framework leverages automated finetuning on synthetically generated, diverse datasets and integrates execution-level testing with mutation-driven unit testing to rigorously verify functional accuracy. Through our evaluations on Java-to-Go and C++-to-Rust translation tasks, ACT demonstrated substantial improvements in translation quality. 

Future work will focus on expanding ACT's functionality and efficiency, supporting other programming languages. We plan to develop a feedback-driven correction loop for test failures to allow for targeted adjustments in both the code and test logic, improving data quality. Additionally, we intend to generate complex, long-form code samples to enhance repository-level code translation tasks.

\bibliography{colm2025_conference}

\begin{thebibliography}{15}
\providecommand{\natexlab}[1]{#1}
\providecommand{\url}[1]{\texttt{#1}}
\expandafter\ifx\csname urlstyle\endcsname\relax
  \providecommand{\doi}[1]{doi: #1}\else
  \providecommand{\doi}{doi: \begingroup \urlstyle{rm}\Url}\fi

\bibitem[Chen et~al.(2021)Chen, Tworek, Jun, Yuan, Pinto, Kaplan, Edwards, Burda, Joseph, Brockman, et~al.]{chen2021evaluating}
Mark Chen, Jerry Tworek, Heewoo Jun, Qiming Yuan, Henrique Ponde De~Oliveira Pinto, Jared Kaplan, Harri Edwards, Yuri Burda, Nicholas Joseph, Greg Brockman, et~al.
\newblock Evaluating large language models trained on code.
\newblock \emph{arXiv preprint arXiv:2107.03374}, 2021.

\bibitem[Dakhel et~al.(2024)Dakhel, Nikanjam, Majdinasab, Khomh, and Desmarais]{dakhel2024effective}
Arghavan~Moradi Dakhel, Amin Nikanjam, Vahid Majdinasab, Foutse Khomh, and Michel~C Desmarais.
\newblock Effective test generation using pre-trained large language models and mutation testing.
\newblock \emph{Information and Software Technology}, 171:\penalty0 107468, 2024.

\bibitem[Hu et~al.(2021)Hu, Shen, Wallis, Allen-Zhu, Li, Wang, Wang, and Chen]{hu2021lora}
Edward~J Hu, Yelong Shen, Phillip Wallis, Zeyuan Allen-Zhu, Yuanzhi Li, Shean Wang, Lu~Wang, and Weizhu Chen.
\newblock Lora: Low-rank adaptation of large language models.
\newblock \emph{arXiv preprint arXiv:2106.09685}, 2021.

\bibitem[Hui et~al.(2024)Hui, Yang, Cui, Yang, Liu, Zhang, Liu, Zhang, Yu, Dang, et~al.]{hui2024qwen2}
Binyuan Hui, Jian Yang, Zeyu Cui, Jiaxi Yang, Dayiheng Liu, Lei Zhang, Tianyu Liu, Jiajun Zhang, Bowen Yu, Kai Dang, et~al.
\newblock Qwen2.5-coder technical report.
\newblock \emph{arXiv preprint arXiv:2409.12186}, 2024.

\bibitem[Karaivanov et~al.(2014)Karaivanov, Raychev, and Vechev]{karaivanov2014phrase}
Svetoslav Karaivanov, Veselin Raychev, and Martin Vechev.
\newblock Phrase-based statistical translation of programming languages.
\newblock In \emph{Proceedings of the 2014 ACM international symposium on new ideas, new paradigms, and reflections on programming \& software}, pp.\  173--184, 2014.

\bibitem[Lu et~al.(2024)Lu, Zhou, Ren, Wang, Shi, Pan, Zhan, and Li]{lu2024mathgenie}
Zimu Lu, Aojun Zhou, Houxing Ren, Ke~Wang, Weikang Shi, Junting Pan, Mingjie Zhan, and Hongsheng Li.
\newblock Mathgenie: Generating synthetic data with question back-translation for enhancing mathematical reasoning of llms.
\newblock \emph{arXiv preprint arXiv:2402.16352}, 2024.

\bibitem[Mukherjee \& Chakrabarti(2011)Mukherjee and Chakrabarti]{mukherjee2011automatic}
Suvam Mukherjee and Tamal Chakrabarti.
\newblock Automatic algorithm specification to source code translation.
\newblock \emph{Indian Journal of Computer Science and Engineering (IJCSE)}, 2\penalty0 (2):\penalty0 146--159, 2011.

\bibitem[Nguyen et~al.(2014)Nguyen, Nguyen, and Nguyen]{nguyen2014migrating}
Anh~Tuan Nguyen, Tung~Thanh Nguyen, and Tien~N Nguyen.
\newblock Migrating code with statistical machine translation.
\newblock In \emph{Companion Proceedings of the 36th International Conference on Software Engineering}, pp.\  544--547, 2014.

\bibitem[Phan \& Jannesari(2020)Phan and Jannesari]{phan2020statistical}
Hung Phan and Ali Jannesari.
\newblock Statistical machine translation outperforms neural machine translation in software engineering: Why and how.
\newblock In \emph{Proceedings of the 1st ACM SIGSOFT International Workshop on Representation Learning for Software Engineering and Program Languages}, pp.\  3--12, 2020.

\bibitem[Rajbhandari et~al.(2020)Rajbhandari, Rasley, Ruwase, and He]{rajbhandari2020zero}
Samyam Rajbhandari, Jeff Rasley, Olatunji Ruwase, and Yuxiong He.
\newblock Zero: Memory optimizations toward training trillion parameter models.
\newblock In \emph{SC20: International Conference for High Performance Computing, Networking, Storage and Analysis}, pp.\  1--16. IEEE, 2020.

\bibitem[Roziere et~al.(2020)Roziere, Lachaux, Chanussot, and Lample]{roziere2020unsupervised}
Baptiste Roziere, Marie-Anne Lachaux, Lowik Chanussot, and Guillaume Lample.
\newblock Unsupervised translation of programming languages.
\newblock \emph{Advances in neural information processing systems}, 33:\penalty0 20601--20611, 2020.

\bibitem[Roziere et~al.(2023)Roziere, Gehring, Gloeckle, Sootla, Gat, Tan, Adi, Liu, Sauvestre, Remez, et~al.]{roziere2023code}
Baptiste Roziere, Jonas Gehring, Fabian Gloeckle, Sten Sootla, Itai Gat, Xiaoqing~Ellen Tan, Yossi Adi, Jingyu Liu, Romain Sauvestre, Tal Remez, et~al.
\newblock Code llama: Open foundation models for code.
\newblock \emph{arXiv preprint arXiv:2308.12950}, 2023.

\bibitem[Weisz et~al.(2022)Weisz, Muller, Ross, Martinez, Houde, Agarwal, Talamadupula, and Richards]{weisz2022better}
Justin~D Weisz, Michael Muller, Steven~I Ross, Fernando Martinez, Stephanie Houde, Mayank Agarwal, Kartik Talamadupula, and John~T Richards.
\newblock Better together? an evaluation of ai-supported code translation.
\newblock In \emph{Proceedings of the 27th International Conference on Intelligent User Interfaces}, pp.\  369--391, 2022.

\bibitem[Xu et~al.(2023)Xu, Sun, Zheng, Geng, Zhao, Feng, Tao, and Jiang]{xu2023wizardlm}
Can Xu, Qingfeng Sun, Kai Zheng, Xiubo Geng, Pu~Zhao, Jiazhan Feng, Chongyang Tao, and Daxin Jiang.
\newblock Wizardlm: Empowering large language models to follow complex instructions.
\newblock \emph{arXiv preprint arXiv:2304.12244}, 2023.

\bibitem[Zhu et~al.(2024)Zhu, Guo, Shao, Yang, Wang, Xu, Wu, Li, Gao, Ma, et~al.]{zhu2024deepseek}
Qihao Zhu, Daya Guo, Zhihong Shao, Dejian Yang, Peiyi Wang, Runxin Xu, Y~Wu, Yukun Li, Huazuo Gao, Shirong Ma, et~al.
\newblock Deepseek-coder-v2: Breaking the barrier of closed-source models in code intelligence.
\newblock \emph{arXiv preprint arXiv:2406.11931}, 2024.

\end{thebibliography}
\bibliographystyle{colm2025_conference}

\appendix
\section{Appendix - Data Generation Working Example}
\label{sec:appendix}

\begin{figure}[hbt!]
    \centering
    \includegraphics[width=\linewidth,trim=0.3cm 0.3cm 0.5cm 2.3cm,clip]{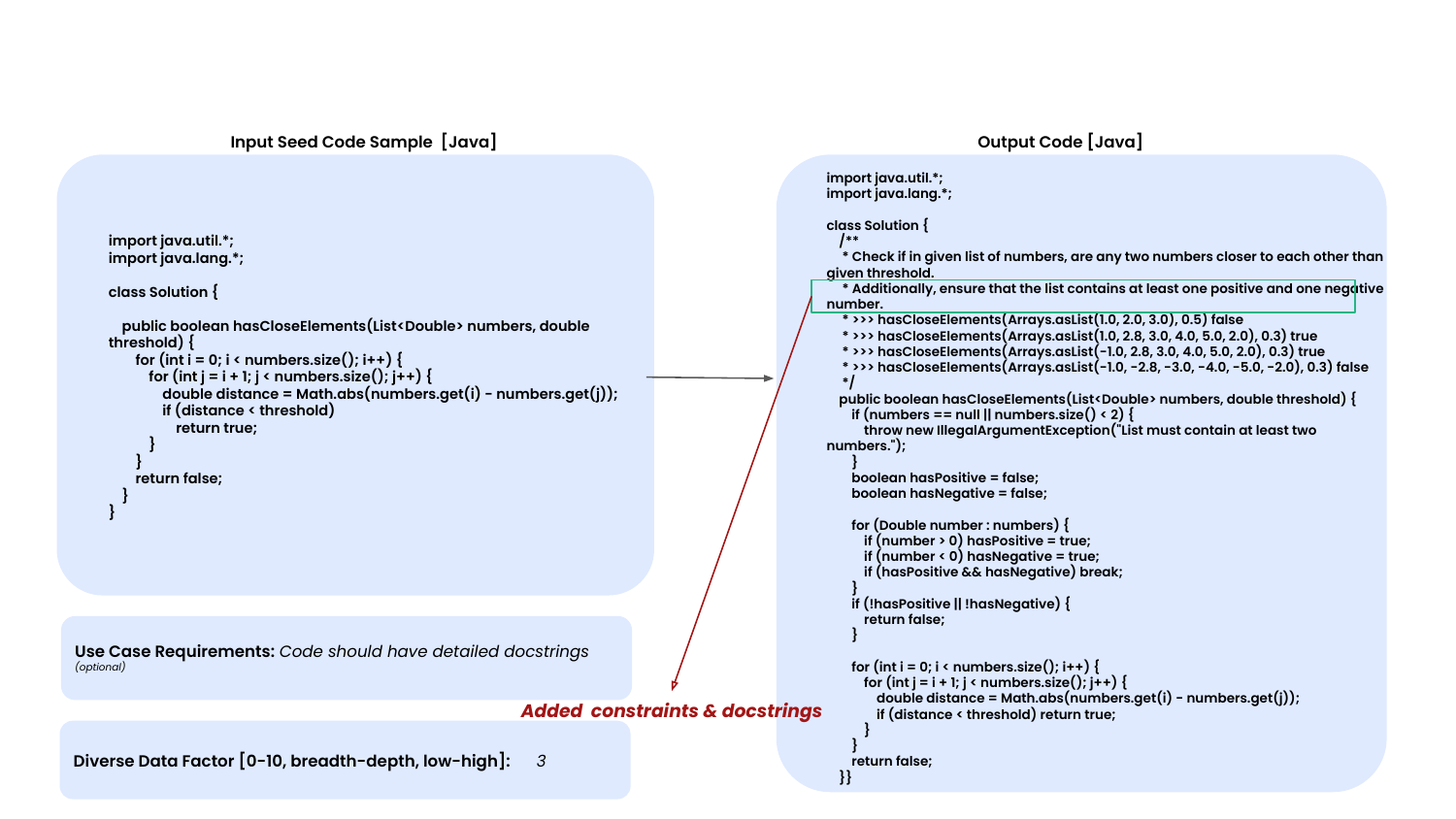}  
    \caption{An example of Source Dataset Depth-based Expansion with Constraint Prompt. 
    The original source code checks if any two numbers in a list are closer than a specified threshold. The dataset expansion pipeline adds a constraint to ensure the list contains at least one positive and one negative number. Additionally, a detailed docstring is included in the evolved code to meet the use case requirements.}
    \label{fig:figure1}
\end{figure}

\begin{figure}[hbt!]
    \centering
    \includegraphics[width=\linewidth,trim=0.3cm 0.3cm 0.5cm 2.3cm,clip]{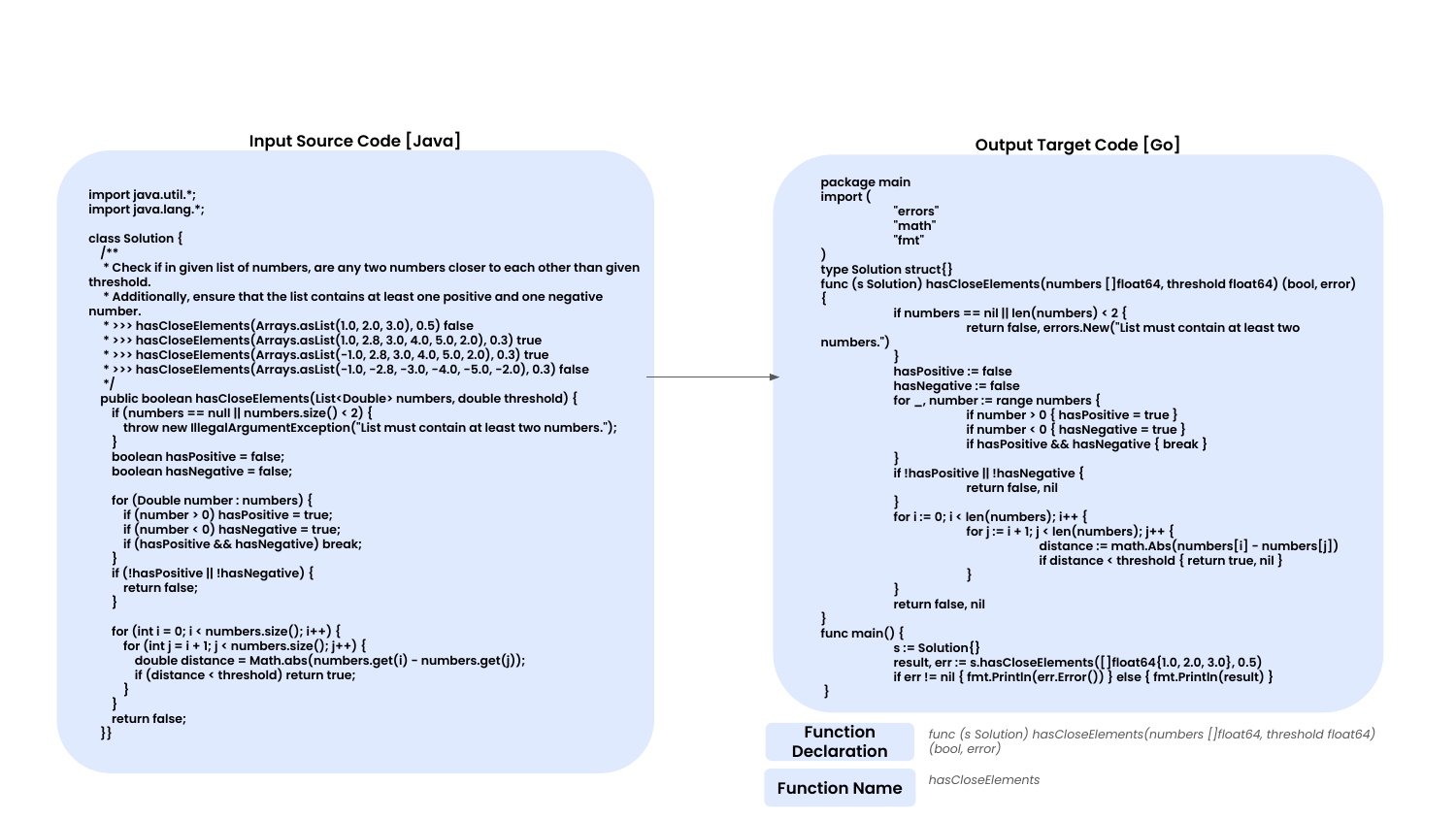}  
    \caption{Translation of source language code into target language code. Function Declaration of the target code is also generated to be used in the next step.}
    \label{fig:figure2}
\end{figure}

\begin{figure}[hbt]
    \centering
    \includegraphics[width=\linewidth,trim=0.3cm 0.1cm 0.5cm 2.3cm,clip]{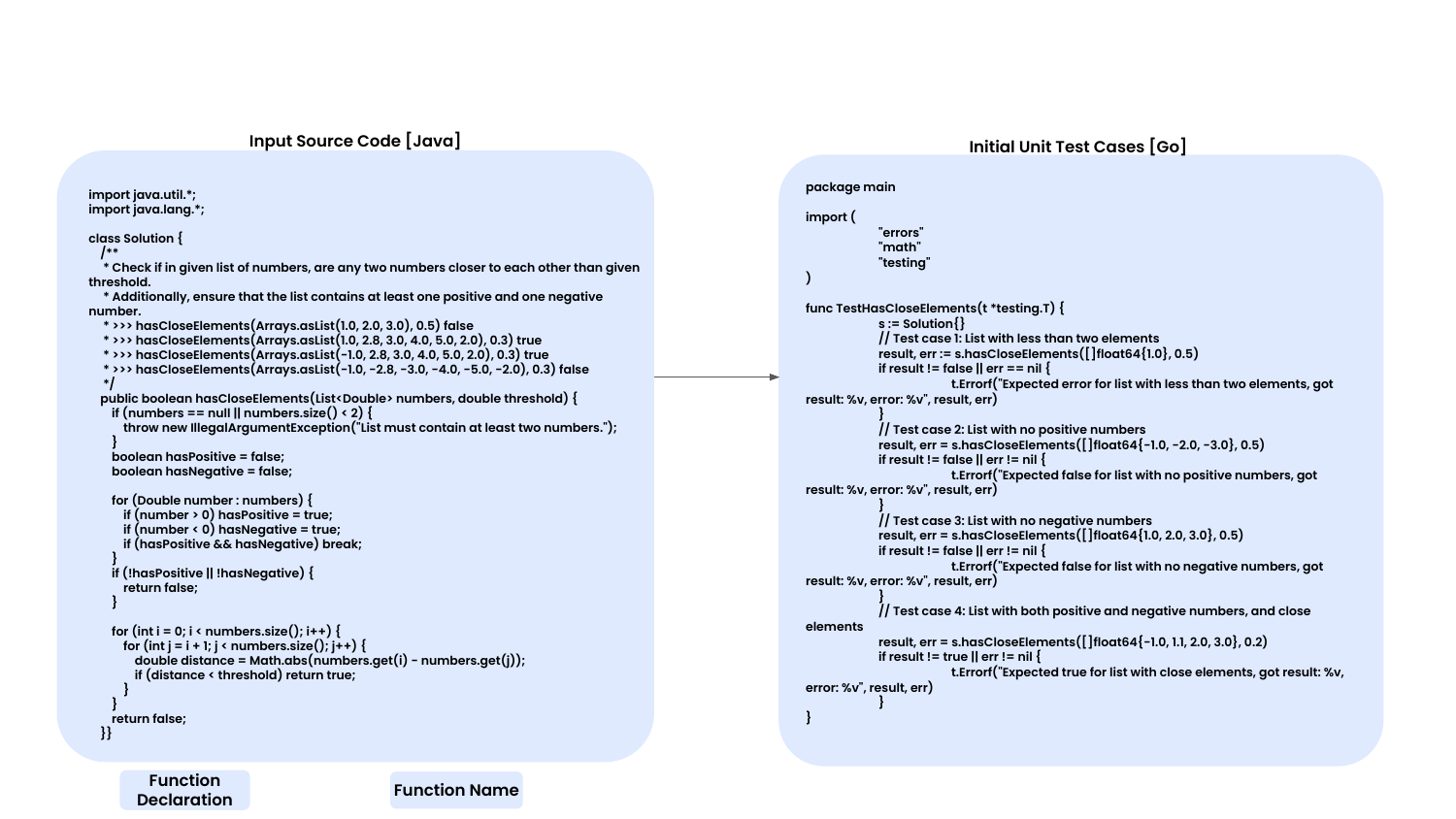}  
    \caption{Generation of initial unit test cases from the source language code (to understand the core logic) and function declaration of the target language code (to understand the function name and parameters used).}
    \label{fig:figure3}
\end{figure}

\begin{figure}[hbt]
    \centering
    \includegraphics[width=\linewidth,trim=0.3cm 0.3cm 0.5cm 2.3cm,clip]{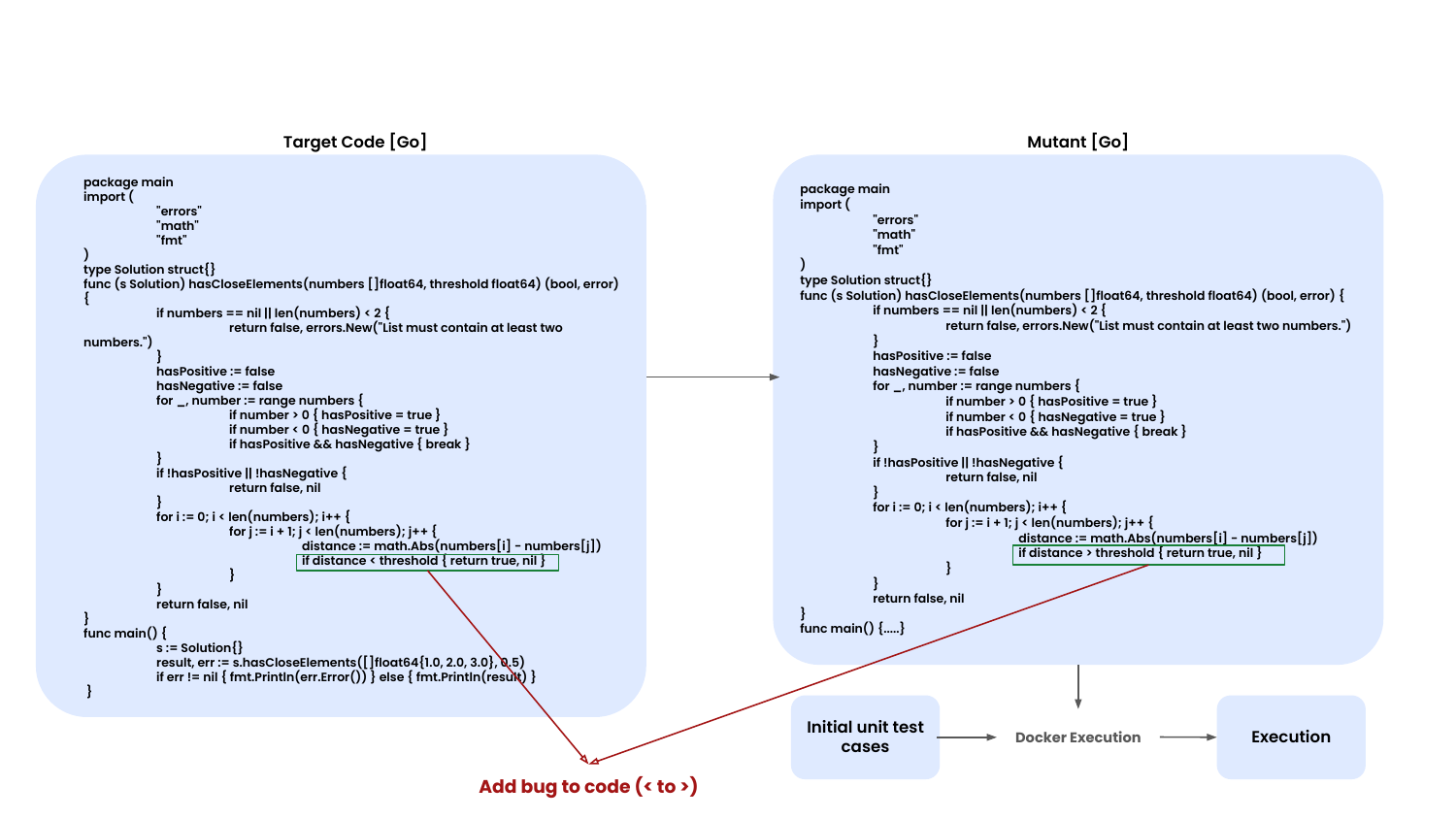}  
    \caption{Mutation Testing of unit test cases. Generation of the mutant code by adding a bug and then verifying if the unit test cases are able to detect the bug.}
    \label{fig:figure4}
\end{figure}

\begin{figure}[hbt]
    \centering
    \includegraphics[width=\linewidth,trim=0.3cm 0.1cm 0.5cm 2.3cm,clip]{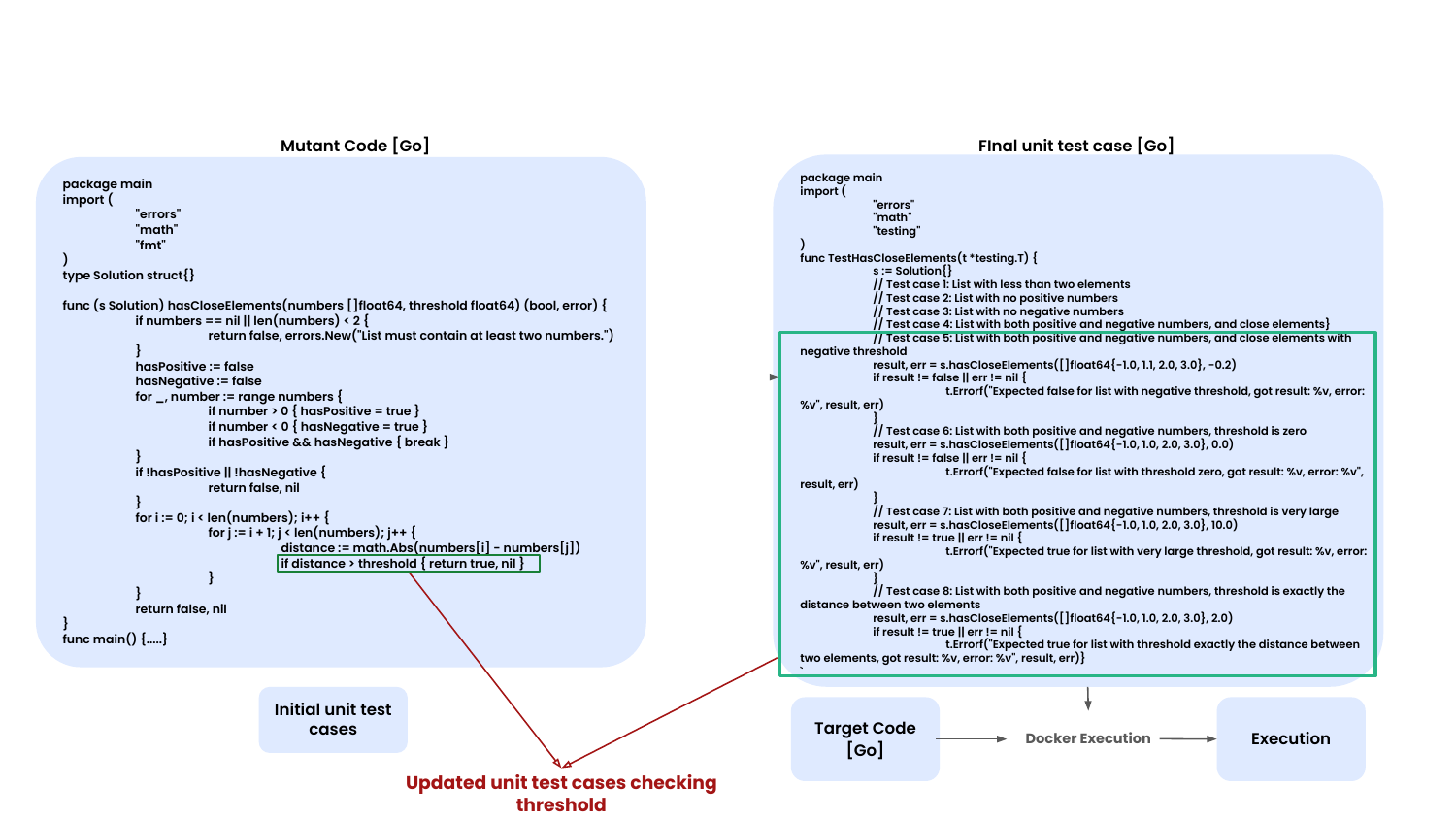}  
    \caption{Final unit test case generation and validation of generated source-target code samples. Initial unit test cases lacked checking the threshold condition. After mutation testing, additional test cases checking various combinations of zero, negative, and large thresholds are generated, leading to better code coverage.}
    \label{fig:figure5}
\end{figure}


\end{document}